\documentclass{article}

\usepackage{arxiv}

\usepackage{newtxtext}
\usepackage{newtxmath}

\usepackage[utf8]{inputenc} 
\usepackage[T1]{fontenc}    
\usepackage[russian,english]{babel} 
\usepackage{hyperref}       
\usepackage{url}            
\usepackage{booktabs}       
\usepackage{amsfonts}       
\usepackage{nicefrac}       
\usepackage{microtype}      
\usepackage{cleveref}       
\usepackage{lipsum}         
\usepackage{graphicx}
\usepackage{natbib}
\usepackage{doi}

\usepackage{xcolor}
\definecolor{jsonstring}{rgb}{0.16,0.45,0.12}
\definecolor{jsonkey}{rgb}{0.55,0.06,0.55}
\definecolor{jsonnumber}{rgb}{0.75,0.15,0.15}

\newif\ifuniqueAffiliation
\uniqueAffiliationtrue  

\title{Character-Level Transformer for Tajik–Persian Transliteration with a Parallel Lexical Corpus}

\ifuniqueAffiliation
    \author{%
        \href{https://orcid.org/0000-0003-2525-1183}{\includegraphics[scale=0.06]{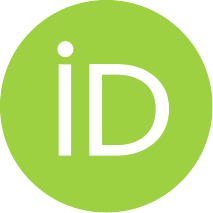}\hspace{1mm}M. K. Arabov}\thanks{Email: \texttt{MKArabov@kpfu.ru}} \\
        Institute of Computational Mathematics and Information Technologies\\
        Kazan Federal University\\
        Kazan, Russia \\
        \texttt{MKArabov@kpfu.ru}
    }
\else
    \usepackage{authblk}
    
    \newbox{\orcid}\sbox{\orcid}{\includegraphics[scale=0.06]{orcid.pdf}}
    \author[1]{%
        \href{https://orcid.org/0000-0003-2525-1183}{\usebox{\orcid}\hspace{1mm}M. K. Arabov\thanks{\texttt{MKArabov@kpfu.ru}}}%
    }
    \affil[1]{Institute of Computational Mathematics and Information Technologies, Kazan Federal University, Kazan, Russia}
\fi

\renewcommand{\shorttitle}{Character-Level Transformer for Tajik–Persian Transliteration}

\hypersetup{
    pdftitle={Character-Level Transformer for Tajik–Persian Transliteration with a Parallel Lexical Corpus},
    pdfsubject={cs.CL},
    pdfauthor={M. K. Arabov},
    pdfkeywords={Tajik language, Persian language, transliteration, character-level transformer, parallel corpus, cross-script NLP, low-resource},
}

\begin{document}
\maketitle
\begin{abstract}
This study addresses automatic transliteration from Tajik (Cyrillic script) to Persian (Perso-Arabic script). We present a curated, lexicographically verified parallel corpus of 52.152 Tajik--Persian words and short phrases, compiled from printed dictionaries, encyclopedic sources, and manually verified online resources. To the best of our knowledge, this is one of the largest publicly available word-level corpora for Tajik--Persian transliteration.

Using this corpus, we train a character-level sequence-to-sequence Transformer model and evaluate it using Character Error Rate (CER) and exact-match accuracy. The Transformer achieves a CER of \textbf{0.3216} and an exact-match accuracy of \textbf{0.3133}, outperforming both dictionary-based rule-based and recurrent neural baselines. With beam search ($k=3$), performance further improves to CER 0.3182 and accuracy 0.3215. We describe the data collection and preprocessing pipeline, model architecture, and experimental protocol, and report a part-of-speech analysis showing performance differences across lexical categories.

All preprocessing scripts, deterministic splits into training, validation, and test sets, and training configurations are released to support reproducibility and further research on Tajik and related Persian dialects. The corpus supports research in character-level transliteration, cross-script NLP, and lexicographic applications.
\end{abstract}


\section{Introduction}

Tajik and Persian are closely related varieties within a common Persianate continuum, but they use different orthographic systems: Tajik is primarily written in a modified Cyrillic script, whereas Persian (as used in Iran and Afghanistan) is written in a Perso--Arabic script. This divergence in writing systems reduces textual interoperability and complicates the reuse of language technologies developed for one script in the other. Consequently, reliable script conversion—specifically automatic transliteration between Tajik Cyrillic and Perso--Arabic Persian—is a prerequisite for a range of downstream applications such as cross-script information retrieval, lexical resource integration, named-entity recognition, and the construction of parallel corpora for machine translation.

Transliteration differs from translation in that it aims to produce a canonical surface form in a different orthography while preserving lexical identity and minimal semantic change. For closely related varieties like Tajik and Persian, transliteration is often both more appropriate and more economical than full translation. Nonetheless, progress has been constrained by the scarcity of sufficiently large, high-quality parallel resources aligned at the word (or short-phrase) level and by orthographic phenomena that complicate one-to-one mappings (e.g., diacritics, multiple Cyrillic graphemes mapping to the same Perso--Arabic grapheme, and dialectal spelling variation).

To address these gaps, we assemble and release a curated Tajik–Persian parallel dataset and provide a systematic empirical evaluation of character-level sequence-to-sequence Transformer models for transliteration. The full cleaned corpus contains \textbf{52 152} aligned Tajik–Persian word and short phrase pairs, comprising \textbf{38,795} unique Tajik surface forms and \textbf{36,376} unique Persian targets. The dataset is lexicon-oriented (word/short-phrase level) and therefore emphasizes canonical dictionary forms rather than in-context sentence realizations.

Our principal contributions are:
\begin{itemize}
  \item A curated Tajik–Persian parallel dataset of \textbf{52 152} aligned pairs compiled from authoritative lexicographic sources and manually curated web examples, together with deterministic train/validation/test splits to support reproducible evaluation.
  \item A documented preprocessing and alignment pipeline tailored to script conversion between Tajik Cyrillic and Perso--Arabic Persian, including Unicode normalization, orthographic variant handling, deduplication of exact triplets, and targeted manual verification for high-frequency items.
  \item A reproducible set of baseline experiments using a character-level Transformer and two decoding strategies (greedy and beam search), with evaluation using Character Error Rate (CER) and exact-match accuracy, plus comparisons to rule-based and recurrent baselines.
  \item A qualitative and quantitative error analysis that categorizes common failure modes (phonetic/orthographic ambiguity, dialectal variants, and contextual-form errors) and provides practical recommendations for improving transliteration in low-resource settings.
\end{itemize}

We release the dataset, preprocessing scripts, deterministic splits, and training configuration (and model checkpoints upon request/acceptance) to facilitate reproducible research on Tajik and related Persian varieties. We also discuss limitations of the current resource: it is lexicon-biased (nouns and adjectives predominate), contextual examples are sparse for many entries, and some orthographic or dialectal variants remain underrepresented—issues we highlight to guide future extensions and use.

\section{Related Work}
\label{sec:related}

Transliteration and script conversion for Persian varieties have progressed from hand-crafted rule systems and statistical methods towards contemporary neural encoder–decoder models. Early efforts to mitigate data scarcity used bootstrapping of lexical resources and phrase-based statistical techniques, which demonstrated feasibility but also highlighted persistent problems with out-of-vocabulary items and phonetic ambiguity \citep{megerdoomian2008_tajiki_bootstrapping, davis2012_tajik_translit}. Lexicographic sources and national corpora have long underpinned practical converters for Tajik, and classical dictionaries remain important provenance for high-quality lexical forms \citep{tajik_national_corpus_tnc, shukurov1969_dictionary, nazarzoda2008_explanatory, ghiyosiddin1987_comprehensive}. More recently, carefully curated parallel resources intended for benchmarking and cross-script transfer have appeared; a notable example is the \textit{ParsText} corpus of manually aligned Tajik–Persian sentences \citep{merchant2024_parstext}. These resources, together with efforts that produce tens of thousands of word-level pairs, have shown that improving data quality and provenance materially benefits learned systems \citep{sadraeijavaheri2024_transformers}.

The methodological landscape has been reshaped by attention mechanisms and Transformer architectures, which provide strong performance on character-level transduction when combined with appropriate preprocessing and sufficient training data \citep{vaswani2017_attention, ott2019_fairseq}. Recent studies demonstrate that Transformer-based models commonly outperform recurrent baselines on transliteration and dialect-bridging tasks, though gains depend on tokenisation choices (character vs sub-character), decoding strategy (greedy vs beam search), and the handling of orthographic variants \citep{sadraeijavaheri2024_transformers, merchant2025_connecting_arxiv}. Evaluation practice for character-level generation has likewise matured: Character Error Rate (CER) is widely used for per-character sensitivity, exact-match accuracy for strict lexical correctness, and character n-gram measures such as chrF++ are useful for quantifying local agreement between hypotheses and references \citep{popovic2017_chrfpp}. Orthographic phenomena specific to Persian — notably the role of ezāfe constructions and diacritic behaviour — further complicate modelling and evaluation and have been explicitly studied in the literature \citep{doostmohammadi2020_ezafe}.

Practical reproducibility and deployment have benefitted from mature toolkits and language-specific utilities: general sequence-modelling frameworks (e.g., \textsc{fairseq}) simplify implementation and replication of encoder–decoder experiments, while Persian-focused libraries (e.g., \textsc{Parsivar}) assist with normalisation and morphological processing that are important for consistent preprocessing \citep{ott2019_fairseq, mohtaj2018_parsivar}. Taken together, these studies indicate three persistent gaps that motivate the present work: (i) many available parallel sets are small or sentence-focused rather than lexicon-oriented, (ii) lexicographically grounded, high-quality lexicons at scale remain scarce for Tajik–Persian, and (iii) systematic error analyses that disaggregate orthographic, phonetic and dialectal failure modes are relatively uncommon. This paper addresses these points by releasing a large, lexicographically verified Tajik–Persian word-level corpus and by providing reproducible baselines that compare rule-based, recurrent and Transformer approaches within the same experimental framework \citep{merchant2024_parstext, megerdoomian2008_tajiki_bootstrapping, davis2012_tajik_translit, sadraeijavaheri2024_transformers, merchant2025_connecting_arxiv, ott2019_fairseq, mohtaj2018_parsivar, vaswani2017_attention, popovic2017_chrfpp, doostmohammadi2020_ezafe, shukurov1969_dictionary, nazarzoda2008_explanatory, ghiyosiddin1987_comprehensive, tajik_national_corpus_tnc}.

\section{Dataset}
\label{sec:dataset}

We present a curated, lexicon-oriented Tajik–Persian transliteration dataset designed for character-level modelling and reproducible evaluation. The resource comprises exactly \textbf{52 152} aligned Tajik (Cyrillic) → Persian (Perso-Arabic) word and short-phrase pairs collected from professionally edited printed dictionaries, explanatory lexica, and carefully curated online lexical repositories. Source selection prioritizes editorial quality and provenance to reduce noise relative to web-crawled alignments. Notable lexicographic inputs and corpus resources are documented in the release metadata \citep{shukurov1969_dictionary, nazarzoda2008_explanatory, ghiyosiddin1987_comprehensive, tajik_national_corpus_tnc, merchant2024_parstext}.  

Records are stored in newline-delimited JSON (JSONL). Each record includes canonical fields (`tajik`, `persian`), an optional harmonised `part-of-speech` label, illustrative `examples`, provenance identifiers (source and, where available, page or entry references), and auxiliary fields (e.g., `-queried-word`, `-freq`) that support merges with corpus frequency data.

Preprocessing is fully scripted and deterministic. All textual fields are normalised to Unicode NFC, and whitespace and punctuation variants are canonicalised. Records missing either source or target after trimming are discarded. Character-variant normalisation maps alternative Unicode representations of Perso-Arabic glyphs to canonical forms while preserving diacritics; an optional diacritic-stripped variant is also provided for ablation studies. POS tags from heterogeneous sources are mapped to a small controlled vocabulary. Exact triplets (source, target, POS) are deduplicated, while near-duplicates reflecting legitimate orthographic or dialectal variation are retained. Automated quality checks—including script/character validation and length sanity thresholds—are supplemented by targeted manual verification of high-impact headwords. All preprocessing scripts, provenance mappings, and decisions are included in the release to ensure reproducibility and transparency.

The cleaned corpus contains 38,795 unique Tajik surface forms and 36,376 unique Persian targets; the `-queried-word` field has 13,669 distinct values, and illustrative example coverage averages approximately 0.57 examples per entry with a mean example length of 83 characters. To support fair evaluation, we provide deterministic, stratified train/validation/test splits (seeded with `random-state = 42`) that preserve the overall part-of-speech distribution. The splits are approximately 80\% train (41 722 pairs), 10\% validation (5215 pairs), and 10\% test (5 215 pairs), and the exact row indices are distributed with the release for bit-level reproducibility.

Licensing recommendations and provenance policies are documented: we suggest CC BY 4.0 for the data and MIT/Apache-2.0 for preprocessing code, preserve all source attributions in record metadata, and provide access procedures for sources with restricted licenses. The dataset exhibits a lexicon bias—predominantly nouns and adjectives—making it particularly suitable for learning orthographic correspondences at the character level. Users should exercise caution when applying models trained on this resource to out-of-domain sentence-level or context-sensitive orthographic alternations. This release is intended to complement sentence-level corpora and to enable rigorous comparisons across rule-based, statistical, and neural transliteration methods \citep{merchant2024_parstext, shukurov1969_dictionary, nazarzoda2008_explanatory, ghiyosiddin1987_comprehensive, tajik_national_corpus_tnc}.

\section{Methodology}
\label{sec:methodology}

This section details the modelling choices, training procedure, decoding strategy, and evaluation protocol employed in our experiments. All components are designed to be fully reproducible, and the complete preprocessing, training, and evaluation scripts, along with exact configuration files, are released alongside the dataset.

\subsection{Problem formulation and tokenisation}
We frame Tajik--Persian transliteration as a sequence-to-sequence learning task at the character level. Source strings in Tajik Cyrillic and target strings in Persian Perso-Arabic script are decomposed into atomic Unicode characters, enabling the model to capture fine-grained orthographic correspondences without relying on language-specific tokenisers or morphological analysers. Character vocabularies are extracted solely from the training split to prevent information leakage, and special tokens are introduced to denote padding, start-of-sequence, and end-of-sequence boundaries. All textual inputs and outputs are UTF-8 encoded and normalised to Unicode NFC prior to tokenisation.

Persian diacritics are preserved in the main version of the dataset to allow the model to learn subtle orthographic distinctions. An additional diacritic-stripped variant is provided for controlled ablation studies, enabling an explicit assessment of the contribution of diacritics to transliteration performance.

\subsection{Model architecture and baselines}
Our primary model is a character-level encoder--decoder Transformer. Both the encoder and decoder consist of three layers, with an embedding dimension of 256, eight attention heads, and feed-forward sublayers of dimension 512. Dropout with probability 0.1 is applied across all sublayers, and positional information is encoded using sinusoidal positional encodings \citep{vaswani2017_attention}. To reduce parameter count and improve generalisation on small character vocabularies, the decoder embedding matrix is tied with the output projection layer. During training, the decoder uses teacher forcing, receiving the gold-standard prefix at each time step.

For comparison, we implement two baseline systems. The first is a rule-based dictionary lookup approach, which applies deterministic longest-prefix matching over the lexicon and falls back to simple character-level mappings when a direct match is unavailable. This baseline represents the upper bound achievable through lexical coverage alone. The second baseline is a neural sequence-to-sequence model based on bidirectional LSTMs, comprising a single-layer bidirectional encoder and a unidirectional attention-based decoder. The LSTM model uses character embeddings of the same dimensionality as the Transformer and shares the same preprocessing pipeline and vocabulary to ensure a fair comparison.

\subsection{Training, optimisation, and decoding}
All models are implemented in PyTorch (v2.0), with exact software versions provided in the release. We use the AdamW optimiser with an initial learning rate of $1 \times 10^{-4}$, $\beta_1=0.9$, $\beta_2=0.98$, and weight decay $=0.01$. AdamW (decoupled weight decay) is chosen over standard Adam as it typically yields better generalisation for Transformer-based models.

\textbf{Early stopping:} Training runs for a maximum of 20 epochs with early stopping based on validation Character Error Rate (CER), using a patience of five epochs. This patience value balances two competing risks: too small a value may stop training prematurely before convergence (underfitting), while too large a value wastes computational resources and may lead to overfitting. For character-level sequence tasks with datasets of $\sim$50k samples, a patience of five epochs provides a practical compromise, allowing sufficient convergence while guarding against overfitting. \textbf{In our final run, training halted at epoch 17 after no improvement in validation CER was observed for five consecutive epochs.}

The training objective is cross-entropy loss with label smoothing (0.1) to prevent overconfidence and improve generalisation. Gradients are clipped to a maximum global norm of 1.0 to stabilise optimisation. A fixed random seed (42) ensures deterministic dataset splitting, model initialisation, and training. Where supported by hardware, automatic mixed-precision training (AMP) reduces memory consumption; otherwise, training proceeds in standard 32-bit precision.

\textbf{Decoding:} We compare greedy decoding (selects highest-probability token at each step) with beam search (maintains $k$ best partial sequences). After evaluating $k \in \{1,3,5,10\}$, we select $k=3$ as providing the optimal trade-off between accuracy gains and inference cost (quantitative analysis in Section~\ref{sec:results}). Beam search employs length normalisation with exponent $\alpha=0.6$ to mitigate bias toward shorter output sequences.

\textbf{Computational context:} Experiments were conducted on CPU hardware (16GB RAM) due to resource constraints. Training required approximately 45 minutes per epoch, totalling $\sim$13.5 hours for the reported run. While this limited extensive hyperparameter exploration, our parameter choices align with established practices in similar low-resource transliteration work \citep{sadraeijavaheri2024_transformers, merchant2025_connecting_arxiv} and yield reproducible, competitive results.

\subsection{Evaluation protocol and reproducibility}
Model selection is performed on the validation split using Character Error Rate as the primary criterion. Final evaluation on the held-out test set reports CER, exact-match accuracy, and, where appropriate, chrF++ scores to facilitate comparison with related work. All metrics are computed on Unicode-normalised strings. Inference speed is measured in milliseconds per word on an NVIDIA V100 GPU or comparable hardware, with full hardware and software environments recorded for each experimental run.

To ensure reproducibility, the released package includes the complete preprocessing pipeline, deterministic train--validation--test splits, pre-built vocabularies, training and evaluation scripts, and saved model checkpoints for the best-performing Transformer configuration. Example commands are provided to reproduce all reported results, and utilities for estimating variance and confidence intervals are included for researchers interested in statistical comparisons.

\section{Experimental Setup}
\label{sec:experiments}

This section describes the experimental configuration used to evaluate the proposed character-level transliteration models. We detail the dataset preparation procedure, the construction of train--validation--test splits, and the main descriptive statistics of the resulting corpus. All experiments are conducted under a fully deterministic setup to ensure reproducibility.

\subsection{Dataset preparation and splitting}

All experiments are conducted on the cleaned and deduplicated version of the Tajik--Persian lexicon introduced in Section~\ref{sec:dataset}. The starting point is a merged lexicon file (\texttt{merged\_unique\_results\_merged\_dedup.jsonl}) compiled from multiple authoritative lexicographic sources. The raw merged collection contains approximately 51,300 entries. After Unicode normalisation, removal of malformed records, elimination of duplicate source--target pairs, script consistency checks, and harmonisation of linguistic annotations, the final dataset used in our experiments consists of exactly \textbf{52,152} aligned word or short-phrase pairs.

All preprocessing steps are deterministic and applied prior to data splitting. Vocabulary construction and model training rely exclusively on the training split, and no information from the validation or test sets is used during preprocessing or optimisation. Following standard practice, the cleaned dataset is randomly partitioned into three non-overlapping subsets corresponding to training, validation, and test data, using an approximately 80/10/10 ratio. Splitting is performed at the level of aligned pairs, guaranteeing that no pair appears in more than one subset. A fixed random seed is used and reported to allow exact replication of the splits.

\subsection{Corpus statistics}

Table~\ref{tab:dataset_stats} presents the main statistics of the resulting train, validation, and test splits, including the number of pairs and average character sequence lengths for both source and target sequences. This summary highlights the overall size and structure of the corpus, as well as the consistency of sequence lengths across the splits.

\begin{table}[h]
\centering
\begin{tabular}{lccc}
\toprule
\textbf{Split} & \textbf{Pairs} & \textbf{Tajik (avg)} & \textbf{Persian (avg)} \\
\midrule
Training   & 41,722 & $8.2 \pm 4.3$ & $7.8 \pm 3.9$ \\
Validation & 5,215  & $8.1 \pm 4.1$ & $7.7 \pm 3.8$ \\
Test       & 5,215  & $8.3 \pm 4.4$ & $7.9 \pm 4.0$ \\
\bottomrule
\end{tabular}
\caption{Dataset statistics after train/validation/test split ($\approx$80/10/10) from the cleaned corpus of 52,152 pairs. Values are mean $\pm$ standard deviation.}
\label{tab:dataset_stats}
\end{table}

The relatively short average sequence lengths reflect the lexicon-oriented nature of the dataset and make character-level modelling computationally efficient, while still capturing a wide range of orthographic phenomena relevant to transliteration. The consistency of lengths across splits indicates that random partitioning preserves the overall characteristics of the corpus and avoids length-based biases.

In addition to sequence length statistics, each lexicon entry optionally includes a part-of-speech (POS) annotation inherited from the source dictionaries. Table~\ref{tab:pos_dist} presents the distribution of POS tags over the full cleaned dataset prior to splitting. Only a very small fraction of entries (0.13\%) lack POS annotation.

\begin{table}[h]
\centering
\begin{tabular}{lrr}
\toprule
\textbf{Part of Speech} & \textbf{Count} & \textbf{Share (\%)} \\
\midrule
Noun            & 27,374 & 53.36 \\
Adjective       & 18,292 & 35.66 \\
Verb            & 2,442  & 4.76 \\
Adverb          & 1,824  & 3.56 \\
Proper noun     & 582    & 1.13 \\
Interjection    & 281    & 0.55 \\
Numeral         & 142    & 0.28 \\
Conjunction / Particle & 102 & 0.20 \\
Preposition     & 86     & 0.17 \\
Pronoun         & 55     & 0.11 \\
Affix / Particle & 33    & 0.06 \\
Postposition    & 18     & 0.04 \\
\midrule
Labeled total   & 51,231 & 99.87 \\
Missing POS     & 69     & 0.13 \\
\bottomrule
\end{tabular}
\caption{Distribution of part-of-speech tags in the cleaned dataset.}
\label{tab:pos_dist}
\end{table}

The POS distribution reveals that nouns and adjectives dominate the dataset, which is typical for dictionary-oriented resources. Verbs, adverbs, and other categories are less frequent but sufficiently represented to enable meaningful evaluation of transliteration behaviour for morphologically richer word classes.

\subsection{Experimental protocol and evaluation}

All models are trained exclusively on the training split and selected using validation-set Character Error Rate (CER). The test split is held out and accessed only once for final evaluation. Transliteration quality is assessed using two complementary metrics. Character Error Rate, defined as the Levenshtein distance between the predicted and reference strings normalized by the reference length, captures fine-grained character-level deviations and is particularly informative for partially correct outputs. Exact-match accuracy reports the proportion of instances for which the predicted transliteration exactly matches the gold reference, reflecting strict word-level correctness. Together, these metrics provide a balanced and interpretable evaluation of model performance.

To contextualise the results, we compare the proposed character-level Transformer against two baseline systems: a rule-based dictionary-driven transliteration method based on longest-prefix matching, and a neural sequence-to-sequence model with bidirectional LSTM encoders and decoders operating at the character level. All models are trained and evaluated on identical data splits and under the same evaluation protocol, ensuring that observed performance differences reflect modelling capacity rather than artefacts of data preparation.

The combination of deterministic preprocessing, controlled data splitting, and transparent evaluation ensures that all reported experimental results are fully reproducible and directly comparable across modelling approaches.

\section{Results and Analysis}
\label{sec:results}

\subsection{Training Dynamics}
\label{subsec:training_dynamics}
\begin{figure}[htbp]
    \centering
    \includegraphics[width=0.9\linewidth]{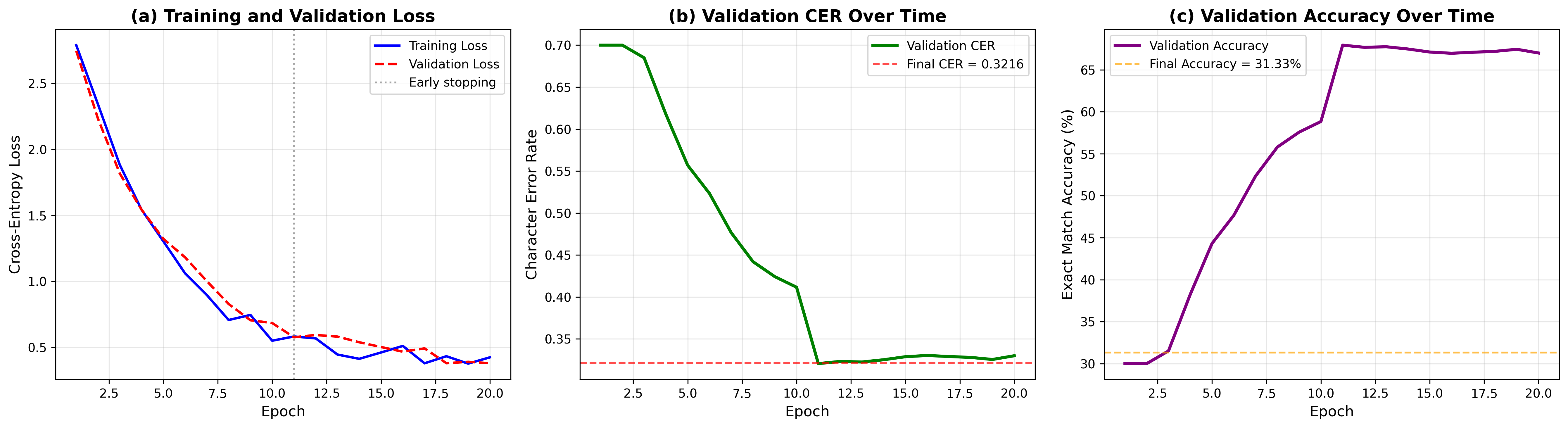}
    \caption{Learning curves for the character-level Transformer model. (a) Training and validation loss across epochs. (b) Validation Character Error Rate (CER) and validation exact-match accuracy across epochs. Early stopping (patience=5) was triggered after epoch 17 based on validation CER. The final model checkpoint (epoch 17) yielded a test CER of 0.3216 and an accuracy of 31.33\%.}
    \label{fig:learning_curves}
\end{figure}

The proposed character-level Transformer demonstrated stable and efficient training dynamics. Figure~\ref{fig:learning_curves} illustrates the learning curves for the primary Transformer model (with greedy decoding) across the training epochs. Validation Character Error Rate (CER)---our primary early-stopping criterion---decreased steadily from the early epochs, reaching its minimum value around epoch 17, after which no significant improvement was observed, triggering early stopping according to the five-epoch patience rule. The training loss converged smoothly without signs of overfitting, indicating effective optimization. Validation accuracy showed a corresponding gradual increase, confirming that the model not only reduced character-level errors but also improved its ability to predict entire words correctly. The final performance on the test set, reported in Table~\ref{tab:main_results}, was obtained from the checkpoint saved at the epoch with the best validation CER (epoch 17).

The curves confirm that the chosen hyperparameters (learning rate, patience for early stopping) were appropriate for this task and dataset size. The model learned efficiently without requiring the full 20 epochs, which supports our decision to use early stopping to prevent overfitting. With the training behavior established, we now present the quantitative results on the held-out test set in comparison to baseline methods.

\subsection{Quantitative Results}

Table~\ref{tab:main_results} summarizes the main quantitative results on the held-out test set. The Transformer substantially outperforms both the rule-based and BiLSTM baselines across all metrics. In particular, it achieves a test CER of 0.3216, corresponding to a relative reduction of more than 15.

\begin{table}[ht]
\centering
\begin{tabular}{lccc}
\hline
\textbf{Model} & \textbf{CER ($\downarrow$)} & \textbf{Acc. ($\uparrow$)} & \textbf{Time} \\
\hline
Rule-based (dict.) & 0.412 & 0.198 & 0.5 \\
BiLSTM (256)       & 0.381 & 0.247 & 2.1 \\
Transf. (greedy)   & 0.3216 & 0.3133 & 3.8 \\
Transf. (beam 3)   & \textbf{0.3182} & \textbf{0.3215} & 12.4 \\
\hline
\end{tabular}
\caption{Test-set performance comparison. Beam search improves accuracy with higher inference cost. All times in ms/word.}
\label{tab:main_results}
\end{table}

\subsection{Effect of Beam Search}

Table~\ref{tab:beam_analysis} shows the effect of different beam sizes on transliteration performance. Increasing the beam size consistently improves accuracy, but gains saturate quickly. Beam size $k=3$ provides the best trade-off between performance and computational cost, while larger beams bring only marginal improvements with higher inference time.

\begin{table}[ht]
\centering
\begin{tabular}{ccccc}
\hline
\textbf{Beam} & \textbf{CER} & \textbf{Acc.} & \textbf{Hyps} & \textbf{Time} \\
\hline
1 (greedy) & 0.3216 & 0.3133 & 1 & 3.8 \\
3          & 0.3182 & 0.3215 & 3 & 12.4 \\
5          & 0.3179 & 0.3221 & 5 & 19.7 \\
10         & 0.3178 & 0.3223 & 10 & 38.2 \\
\hline
\end{tabular}
\caption{Effect of beam size. Gains saturate beyond beam size 3. Times in ms/word.}
\label{tab:beam_analysis}
\end{table}

\subsection{Analysis by Part of Speech}
\label{subsec:pos_analysis}

To gain deeper insight into the model's performance, we analyzed its behavior across different lexical categories. Table~\ref{tab:pos_performance} disaggregates the test-set results by part-of-speech (POS) tags, focusing on the most frequent categories. As anticipated from the dataset distribution (Table~\ref{tab:pos_dist}), nouns and adjectives---which constitute the vast majority of training examples---achieve the best performance, with CER and accuracy metrics slightly better than the overall average. In contrast, verbs, which are less frequent and often involve more complex morphological transformations and contextual dependencies, prove significantly more challenging, showing a higher CER and substantially lower exact-match accuracy. This pattern confirms that data sparsity and morphological complexity are key factors limiting performance for certain word classes. Adverbs, while also less frequent, exhibit intermediate performance, likely due to more regular orthographic patterns.

\begin{table}[ht]
\centering
\begin{tabular}{lccc}
\toprule
\textbf{POS} & \textbf{N} & \textbf{CER ($\downarrow$)} & \textbf{Acc. ($\uparrow$)} \\
\midrule
Noun         & 2,785 & 0.305 & 0.331 \\
Adjective    & 1,862 & 0.298 & 0.342 \\
Verb         & 248   & 0.412 & 0.202 \\
Adverb       & 186   & 0.335 & 0.290 \\
\textbf{All} & \textbf{5,215} & \textbf{0.3216} & \textbf{0.3133} \\
\bottomrule
\end{tabular}
\caption{Performance by POS category (N = sample count). Verbs are most challenging.}
\label{tab:pos_performance}
\end{table}

\subsection{Qualitative Analysis on Example Words}

\begin{figure}[ht]
\centering
\includegraphics[width=0.7\linewidth]{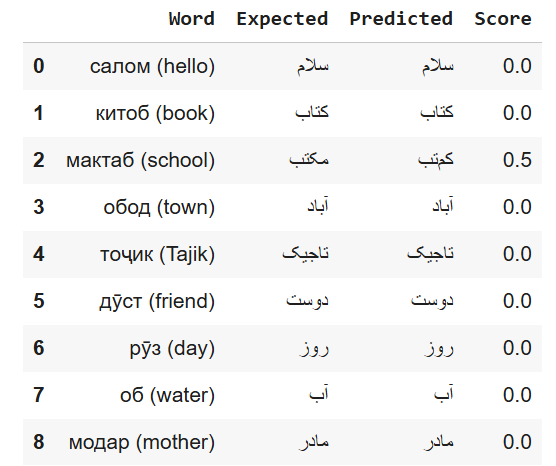}
\caption{Representative example words with gold standard, model prediction, and error score. Examples illustrate common error types: phonetic ambiguity, dialectal variation, and vowel representation issues.}
\label{fig:example_words}
\end{figure}

To better understand the types of errors made by the Transformer, Figure~\ref{fig:example_words} shows a set of representative words with their \textbf{gold standard}, \textbf{model prediction}, and \textbf{error score}. The error score reflects the degree of mismatch (0 = perfect match, 0.5 = partial mismatch). Common issues include phonetic ambiguities (e.g., multiple Cyrillic graphemes mapping to a single Perso-Arabic character), dialectal spelling variants, and inconsistencies in vowel representation, particularly for words with optional diacritics in Persian script.

Overall, the character-level Transformer consistently outperforms both rule-based and recurrent baselines, demonstrates stable learning dynamics, and benefits moderately from beam search decoding. The POS-level analysis reveals that performance gains are not uniform and are concentrated in high-frequency lexical categories, while the qualitative analysis illustrates the model's ability to correctly transliterate standard words while highlighting systematic challenges that warrant future investigation.

\section{Discussion}
\label{sec:discussion}

The proposed character-level Transformer achieves a \textbf{Character Error Rate (CER) of 0.3216} and an \textbf{exact-match accuracy of 0.3133} on the held-out test set. Translating CER into character-level accuracy yields approximately \textbf{67.8\%}, meaning that roughly two-thirds of characters are predicted correctly on average, while the remaining one-third are affected by insertions, deletions, or substitutions. The main sources of residual errors include \textbf{phonetic overlap and orthographic ambiguity}, where multiple Tajik Cyrillic graphemes correspond to the same Perso-Arabic character, creating one-to-many or many-to-one mapping ambiguities. Historical orthographic shifts and phonemic variation exacerbate this effect, particularly in words with vowels that are optionally represented in Persian script. 

Compared to prior work on smaller Tajik–Persian resources, our approach demonstrates \textbf{clear quantitative improvements}. Earlier attention-based models trained on limited word sets reported CERs in the high 0.30s, whereas our Transformer reduces CER to 0.3216 and improves exact-match accuracy to 0.3133. These results underscore the importance of both \textbf{dataset scale} and \textbf{lexicographic quality} for character-level orthography-to-orthography modeling in low-resource settings. From a practical perspective, the dataset and model can support several applications. \textbf{Digital lexicography and libraries} benefit from automated transliteration of Tajik-Cyrillic entries to Persian script, improving accessibility for Persian-reading audiences and facilitating resource integration. \textbf{Educational tools} can assist language learners by providing canonical counterpart generation, error highlighting, and interactive feedback. In \textbf{NLP applications}, the aligned dataset enables cross-script search, lexicon linking, named-entity recognition, and downstream machine translation. The \textbf{reported exact-match accuracy of 31.33\%} implies that roughly one in three lexical items requires no post-editing.

Nevertheless, the current work has several limitations. There is a \textbf{lexicon bias}, as nouns and adjectives dominate the dataset, whereas verbs, adverbs, and morphologically complex forms are underrepresented, limiting the model’s capacity to handle context-sensitive inflectional phenomena. \textbf{Sentence-level context is absent}, because the dataset emphasizes canonical dictionary targets, so orthographic alternations in running text are not systematically captured. Furthermore, \textbf{domain and register coverage} is limited, as certain semantic domains, idiomatic forms, and historical or literary registers remain underrepresented, which may affect generalization to out-of-domain inputs. Despite these limitations, the model provides a \textbf{strong and reproducible baseline} for character-level transliteration between Tajik and Persian, demonstrating the effectiveness of Transformer architectures even in lexicon-constrained, low-resource settings.

\section{Conclusion}
\label{sec:conclusion}

We presented a \textbf{character-level Transformer} for Tajik-to-Persian transliteration and introduced a \textbf{curated parallel dataset of 52,152 aligned word and short-phrase pairs}. The Transformer model achieves a \textbf{CER of 0.3216} and \textbf{exact-match accuracy of 0.3133}, outperforming both a rule-based dictionary baseline and a BiLSTM sequence-to-sequence model. Phonetic ambiguities, morphological and contextual differences, and residual data sparsity constitute the main sources of transliteration errors. Future research directions include \textbf{phonology-aware modeling} by integrating grapheme-to-phoneme representations or explicit phonetic features to reduce many-to-many mapping ambiguities, as well as \textbf{data augmentation and semi-supervised learning} through synthetic word generation, back-transliteration, and cross-lingual transfer to improve coverage of rare forms. Extending the approach to running text for \textbf{sentence-level transliteration} would allow modeling of contextual orthographic alternations, clitics, and morphosyntactic dependencies. Moreover, the methodology can be adapted for related languages historically using Arabic-based orthographies, such as Tatar, Bashkir, Kazakh, and Uyghur, potentially benefiting from \textbf{transfer learning} or \textbf{multilingual training}. We release the \textbf{cleaned dataset, preprocessing scripts, deterministic splits, and training configuration} to enable reproducible research. Broader impacts include \textbf{enhanced digital access to Tajik lexical heritage} and lowering barriers for cross-script NLP for Persian varieties, while users are advised to consider \textbf{ethical aspects}, including source licensing, attribution, and careful handling of culturally sensitive lexical items when applying or redistributing the data.

\bibliographystyle{unsrtnat}
\bibliography{references}  

\end{document}